\documentclass[11pt,a4paper]{article}
\usepackage[hyperref]{emnlp2018}
\usepackage{times}
\usepackage{latexsym}

\usepackage{amsmath}
\usepackage{multirow}
\usepackage{comment}
\usepackage{amssymb}
\usepackage{amsmath}
\usepackage{booktabs}
\usepackage{graphicx}
\usepackage{rotating}
\usepackage{setspace}
\usepackage{caption}

\usepackage{url}

\aclfinalcopy 

\title{On the Role of Text Preprocessing in Neural Network Architectures: \\ An Evaluation Study on Text Categorization and Sentiment Analysis}

\author{Jose Camacho-Collados \\
  School of Computer Science \\ and Informatics\\
  Cardiff University\\
  {\tt camachocolladosj@cardiff.ac.uk} \\
  \And
  Mohammad Taher Pilehvar\\
  School of Computer Engineering\\
  Iran University of \\Science and Technology\\
  {\tt pilehvar@iust.ac.ir} \\}

\date{}

\begin{document}






\maketitle
\begin{abstract}
 Text preprocessing is often the first step in the pipeline of a Natural Language Processing (NLP) system, with potential impact in its final performance.
 Despite its importance, text preprocessing has not received much attention in the deep learning literature.
 In this paper we investigate the impact of simple text preprocessing decisions (particularly tokenizing, lemmatizing, lowercasing and multiword grouping) on the performance of a standard neural text classifier. 
 We perform an extensive evaluation on standard benchmarks from text categorization and sentiment analysis. 
 While our experiments show that a simple tokenization of input text is generally adequate, they also highlight significant degrees of variability across preprocessing techniques. This reveals the importance of paying attention to this usually-overlooked step in the pipeline, particularly when comparing different models. Finally, our evaluation provides insights into the best preprocessing practices for training word embeddings.
\end{abstract}


\section{Introduction}

Words are often considered as the basic constituents of texts for many languages, including English.\footnote{Note that although word-based models are mainstream in NLP in general and text classification in particular, recent work has also considered other linguistic units, such as characters  \cite{kim2015character,XiaoCho:2016} or word senses \cite{LiJurafsky:2015,flekovasupersense,pilehvaracl17}. These techniques require a different kind of preprocessing and, while they have been shown effective in various settings, in this work we only focus on the mainstream word-based models.}
The first module in an NLP pipeline is a tokenizer which transforms texts to sequences of words.
However, in practise, other preprocessing techniques can be (and are) further used together with tokenization. 
These include lemmatization, lowercasing and multiword grouping, among others.
Although these preprocessing decisions have been studied in the context of conventional text classification techniques \cite{leopold2002text,uysal2014impact}, little attention has been paid to them in the more recent neural-based models. The most similar study to ours is \newcite{zhang2017encoding}, which analyzed different encoding levels for English and Asian languages such as Chinese, Japanese and Korean. As opposed to our work, their analysis was focused on UTF-8 bytes, characters, words, romanized characters and romanized words as encoding levels, rather than the preprocessing techniques analyzed in this paper. 

Additionally, word embeddings have been shown to play an important role in boosting the generalization capabilities of neural systems \cite{goldberg2016primer,CamachoColladosPilehvar:2018survey}
. However, while some studies have focused on intrinsically analyzing the role of lemmatization in their underlying training corpus \cite{ebert2016lamb,kuznetsov2018text}, the impact on their extrinsic performance when integrated into a neural network architecture has remained understudied.\footnote{Not only the preprocessing of the corpus may play an important role but also its nature, domain, etc. \newcite{levy2015improving} also showed how small hyperparameter variations may have an impact on the performance of word embeddings. However, these considerations remain out of the scope of this paper.}

In this paper we focus on the role of preprocessing the input text, particularly in how it is split into individual (meaning-bearing) tokens and how it affects the performance of standard neural text classification models based on Convolutional Neural Networks \cite[CNN]{lecun2010convolutional,kim2014convolutional}. 
CNNs have proven to be effective in a wide range of NLP applications, including text classification tasks such as topic categorization \cite{johnson2015effective,tang2015document,XiaoCho:2016,conneau-EtAl:2017:EACLlong} and polarity detection \cite{kalchbrenner2014convolutional,kim2014convolutional,dos2014deep,yin2017comparative}, which are the tasks considered in this work. 
The goal of our evaluation study is to find answers to the following two questions:%
\begin{enumerate}
\item Are neural network architectures (in particular CNNs) affected by seemingly small preprocessing decisions in the input text?     
\item Does the preprocessing of the embeddings' underlying training corpus have an impact on the final performance of a state-of-the-art neural network text classifier?
\end{enumerate}


According to our experiments in topic categorization and polarity detection, these decisions are important in certain cases. 
Moreover, we shed some light on the motivations of each preprocessing decision and provide some hints on how to normalize the input corpus to better suit each setting.

The accompanying materials of this submission can be downloaded at the following repository: \url{github.com/pedrada88/preproc-textclassification}.

\section{Text Preprocessing}
\label{preprocessing}

Given an input text, words are gathered as input units of classification models through tokenization. 
We refer to the corpus which is only tokenized as \textit{\textbf{vanilla}}. For example, given the sentence ``Apple is asking its manufacturers to move MacBook Air production to the United States.'' (running example), the vanilla tokenized text would be as follows (white spaces delimiting different word units): 

\paragraph{} 
\textit{Apple is asking its manufacturers to move MacBook Air production to the United States .}
\newline

We additionally consider three simple preprocessing techniques to be applied to an input text: lowercasing (Section \ref{lowercasing}), lemmatizing (Section \ref{lemmatizing}) and multiword grouping (Section \ref{multiwords}).

\subsection{Lowercasing}
\label{lowercasing}

This is the simplest preprocessing technique which consists of lowercasing each single token of the input text: 

\paragraph{} 
\textit{apple is asking its manufacturers to move macbook air production to the united states .}
\newline

Due to its simplicity, lowercasing has been a popular practice in modules 
of deep learning libraries 
and word embedding packages \cite{pennington2014glove,faruqui2014retrofitting}. 
Despite its desirable property of reducing sparsity and vocabulary size, lowercasing may negatively impact system's performance by increasing ambiguity. For instance, the \textit{Apple} company in our example and the \textit{apple} fruit would be considered as identical entities. 

\subsection{Lemmatizing}
\label{lemmatizing}

The process of lemmatizing consists of replacing a given token with its corresponding lemma: 

\paragraph{} 
\textit{Apple be ask its manufacturer to move MacBook Air production to the United States .}
\newline

Lemmatization has been traditionally a standard preprocessing technique for linear text classification systems \cite{mullen2004sentiment,toman2006influence,hassan2007random}. However, it is rarely used as a preprocessing stage in neural-based systems. The main idea behind lemmatization is to reduce sparsity, as different inflected forms of the same lemma may occur infrequently (or not at all) during training. However, this may come at the cost of neglecting important syntactic nuances.

\subsection{Multiword grouping}
\label{multiwords}

This last preprocessing technique consists of grouping consecutive tokens together into a single token if found in a given inventory:

\paragraph{} 
\textit{Apple is asking its manufacturers to move MacBook\_Air production to the United\_States .}
\newline

The motivation behind this step lies in the idiosyncratic nature of multiword expressions \cite{sag2002multiword}, e.g. \textit{United States} in the example. The meaning of these multiword expressions are often hardly traceable from their individual tokens. As a result, treating multiwords as single units may lead to better training of a given model. Because of this, word embedding toolkits such as Word2vec propose statistical approaches for extracting these multiwords, or directly include multiwords along with single words in their pre-trained embedding spaces \cite{mikolov2013distributed}.


\section{Evaluation}

We considered two tasks for our experiments: \textbf{topic categorization}, i.e. assigning a topic to a given document from a pre-defined set of topics, and \textbf{polarity detection}, i.e. detecting if the sentiment of a given piece of text is positive or negative \cite{dong2015statistical}. 
Two different settings were studied: (1) word embedding's training corpus and the evaluation dataset were preprocessed in a similar manner (Section \ref{experiment1}); and (2) the two were preprocessed differently (Section \ref{experiment2}). In what follows we describe the common experimental setting as well as the datasets and preprocessing used for the evaluation.



\subsection{Experimental setup} 

We tried with two classification models.
The first one is a standard CNN model similar to that of \newcite{kim2014convolutional}, using ReLU \cite{icml2010_NairH10} as non-linear activation function. In the second model, we add a recurrent layer (specifically an LSTM \cite{hochreiter1997long}) before passing the pooled features directly to the fully connected softmax layer.\footnote{The code for this CNN implementation is the same as in \cite{pilehvaracl17}, which is available at \url{https://github.com/pilehvar/sensecnn} } The inclusion of this LSTM layer has been shown to be able to effectively replace multiple layers of convolution and be beneficial particularly for large inputs \cite{XiaoCho:2016}. These models were used for both topic categorization and polarity detection tasks, with slight hyperparameter variations given their different natures (mainly in their text size) which were fixed across all datasets. 
The embedding layer was initialized using 300-dimensional CBOW Word2vec embeddings \cite{Mikolovetal:2013} trained on the 3B-word UMBC WebBase corpus \cite{han2013umbc} with standard hyperparameters\footnote{Context window of 5 words and hierarchical softmax.}. 

\paragraph{Evaluation datasets.} 
For the topic categorization task we used the \textbf{BBC} news dataset\footnote{\url{http://mlg.ucd.ie/datasets/bbc.html}} 
\cite{greene2006practical}, \textbf{20News} \cite{lang1995newsweeder}, \textbf{Reuters}\footnote{Due to the large number of labels in the original Reuters (i.e. 91) and to be consistent with the other datasets, we reduce the dataset to its 8 most frequent labels, a reduction already performed in previous works \cite{sebastiani2002machine}.} 
\cite{lewis2004rcv1} and \textbf{Ohsumed}\footnote{\url{ftp://medir.ohsu.edu/pub/ohsumed}}. 
\textbf{PL04} \cite{Pang+Lee:04a}, \textbf{PL05}\footnote{Both PL04 and PL05 were downloaded from \url{http://www.cs.cornell.edu/people/pabo/movie-review-data/}}
\cite{Pang+Lee:05a}, \textbf{RTC}\footnote{\url{http://www.rottentomatoes.com}}, \textbf{IMDB} \cite{maas-EtAl:2011:ACL-HLT2011} and the Stanford sentiment dataset\footnote{We mapped the numerical value of phrases to either negative (from 0 to 0.4) or positive (from 0.6 to 1), removing the neutral phrases according to the scale (from 0.4 to 0.6).} \cite[\textbf{SF}]{SocherEtAl2013:RNTN}
were considered for polarity detection. 
Statistics of the versions of the datasets used are displayed in Table \ref{tab:statsdatasets}.\footnote{For the datasets with train-test partitions, the sizes of the test sets are the following: 7,532 for 20News; 12,733 for Ohsumed; 25,000 for IMDb; and 1,000 for RTC.} For both tasks the evaluation was carried out either by 10-fold cross-validation or using the train-test splits of the datasets, in case of availability.

\begin{table}
\begin{center}
{
\setlength{\tabcolsep}{6.0pt}
\scalebox{0.78}{ 
\begin{tabular}{l|llrrl}
\toprule
&\bf Dataset  &
\bf Type &
\bf Labels&
\bf \# of docs &
\bf Eval. \\

\midrule
\multirow{4}{*}{\begin{sideways}\bf TOPIC\end{sideways}} & BBC	& News &  5  &  2,225   
&  10-cross  \\
 & 20News & News &  6  &  18,846   
& Train-test  \\
& Reuters & News &  8  &  9,178     & 10-cross     \\
 & Ohsumed & Medical &   23  &  23,166   & Train-test    \\

\midrule

\multirow{5}{*}{\begin{sideways}\bf POLARITY \end{sideways}}  & RTC & Snippets & 2  &  438,000    & Train-test  \\
 & IMDB	& Reviews  & 2  &  50,000    & Train-test  \\
 & PL05 & Snippets & 2  &  10,662      & 10-cross     \\
 & PL04 & Reviews & 2   &  2,000     & 10-cross     \\
 &  Stanford & Phrases & 2  &  119,783    & 10-cross     \\
\bottomrule

\end{tabular}
}
}
\end{center}
\caption{\label{tab:statsdatasets} 
Evaluation datasets for topic categorization and polarity detection.}
\end{table}

\paragraph{Preprocessing.} 
Four different techniques (see Section \ref{preprocessing}) were used to preprocess the datasets as well as the corpus which was used to train word embeddings (i.e. UMBC). For tokenization and lemmatization we relied on Stanford CoreNLP \cite{manning-EtAl:2014:P14-5}. As for multiwords, we used the phrases from the pre-trained Google News Word2vec vectors, which were obtained using a simple statistical approach 
\cite{mikolov2013distributed}.\footnote{For future work it would be interesting to explore more complex methods to learn embeddings for multiword expressions \cite{yin2014exploration,Poliak:2017EACL}.} 


\begin{table*}[t]
\begin{center}
\setlength{\tabcolsep}{7.5pt}
\scalebox{0.90}{
      \begin{tabular}{c c l c  c  c  c c c c c c}
        \toprule
     &   
   \multicolumn{1}{c}{}
   &  \multicolumn{4}{c}{\textbf{Topic categorization}}  &  \multicolumn{5}{c}{\textbf{Polarity detection}}  \\

       \cmidrule(lr){3-6}
      \cmidrule(lr){7-11}

      
&     \multicolumn{1}{l}{\small \textbf{Preprocessing}} &   \multicolumn{1}{c}{\textbf{BBC}} & \multicolumn{1}{c}{\textbf{20News}} & \multicolumn{1}{c}{\textbf{Reuters}} & \multicolumn{1}{c}{\textbf{Ohsumed}} & \multicolumn{1}{c}{\textbf{RTC}} &
      \multicolumn{1}{c}{\textbf{IMDB}} &
      \multicolumn{1}{c}{\textbf{PL05}} &  \multicolumn{1}{c}{\textbf{PL04}} & 
      \multicolumn{1}{c}{\textbf{SF}}\\
      
    \midrule
    
       
       
 \multirow{4}{*}{\rotatebox[origin=c]{90}{\small \textbf{CNN}}}  & \multicolumn{1}{|l}{\text{Vanilla}} &  94.6 & 89.2 & 93.7 & 35.3 & \bf 83.2 & 87.5 & 76.3 & ~~58.7$^\dagger$ & \bf 91.2 \\
     &   \multicolumn{1}{|l}{\text{Lowercased}}& 94.8 & \bf 89.8 & \bf 94.2   & \bf 36.0 & 83.0  & ~~84.2$^\dagger$ & 76.1 & ~~59.6$^\dagger$ & 91.1 \\
     &  \multicolumn{1}{|l}{\text{Lemmatized}} & 95.4 & 89.4 & 94.0   & 35.9 & 83.1  & ~~86.8$^\dagger$ & ~~75.8$^\dagger$ & \bf 64.2 & \bf 91.2 \\
    &   \multicolumn{1}{|l}{\text{Multiword}}  & \bf 95.5 & 89.6 & ~~93.4$^\dagger$  & ~~34.3$^\dagger$ & \bf 83.2  & \bf 87.9 & \bf  77.0 & ~~59.1$^\dagger$ & \bf 91.2 \\

\midrule
\midrule

 \multirow{4}{*}{\rotatebox[origin=c]{90}{\small \textbf{CNN+LSTM}}}  &     \multicolumn{1}{|l}{\text{Vanilla}} & \bf 97.0  & 90.7 & 93.1 & ~~30.8$^\dagger$ & \bf 84.8  & \bf 88.9 & 79.1 & 71.4 & 87.1 \\
     &   \multicolumn{1}{|l}{\text{Lowercased}} & 96.4  & \bf 90.9 & 93.0 & \bf 37.5 & 84.0  & ~~88.3$^\dagger$ & \bf 79.5 & \bf 73.3 & 87.1 \\
 \multicolumn{1}{c}{\textbf{}}    &  \multicolumn{1}{|l}{\text{Lemmatized}} & 95.8$^\dagger$  & 90.5 & \bf 93.2 & 37.1 & 84.4  & ~~87.7$^\dagger$ & 78.7 & 72.6 & 86.8$^\dagger$ \\
 \multicolumn{1}{c}{\textbf{}}   &   \multicolumn{1}{|l}{\text{Multiword}} & 96.2  & ~~89.8$^\dagger$ & ~~92.7$^\dagger$ & ~~29.0$^\dagger$ & 84.0  & \bf 88.9 & 79.2 & ~~67.0$^\dagger$ & \bf 87.3 \\

       \bottomrule
       
    \end{tabular}
    }
    \end{center}
    \caption{Accuracy on the topic categorization and polarity detection tasks using various preprocessing techniques for the CNN and CNN+LSTM models. $^\dagger$ indicates results that are statistically significant with respect to the top result. 
    }
    \bigskip
    \label{tab:results-experiment1}
\end{table*}




\subsection{Experiment 1: Preprocessing effect}
\label{experiment1}

Table \ref{tab:results-experiment1} shows the accuracy\footnote{Computed by averaging accuracy of two different runs. The statistical significance was calculated according to an unpaired t-test at the 5\% significance level.} of the classification models using our four preprocessing techniques. 
We observe a certain variability of results depending on the preprocessing techniques used (average variability\footnote{Average variability was the result of averaging the variability of each dataset, which was computed as the difference between the best and the worst preprocessing performances.} of $\pm 2.4\%$ for the CNN+LSTM model, including a statistical significance gap in seven of the nine datasets), which proves the influence of preprocessing on the final results. It is perhaps not surprising that the lowest variance of results is seen in the datasets with the larger training data (i.e. RTC and Stanford). This suggests that the preprocessing decisions are not so important when the training data is large enough, but they are indeed relevant in benchmarks where the training data is limited.

As far as the individual preprocessing techniques are concerned, the vanilla setting (tokenization only) proves to be consistent across datasets and tasks, as it performs in the same ballpark as the best result in 8 of the 9 datasets for both models (with no noticeable differences between topic categorization and polarity detection). The only topic categorization dataset in which tokenization does not seem enough is Ohsumed, which, unlike the more general nature of 
other categorization datasets (news), belongs to a specialized domain (medical) for which 
fine-grained distinctions are required to classify cardiovascular diseases. 
In particular for this dataset, word embeddings trained on a general-domain corpus like UMBC may not accurately capture the specialized meaning of medical terms and hence, sparsity becomes an issue.
In fact, lowercasing and lemmatizing, which are mainly aimed at reducing sparsity, outperform the vanilla setting by over six points in the CNN+LSTM setting and clearly outperform the other preprocessing techniques on the single CNN model as well. 

Nevertheless, the use of more complex preprocessing techniques such as lemmatization and multiword grouping does not help in general. Even though lemmatization has 
proved useful in conventional linear models as an effective way to deal with sparsity \cite{mullen2004sentiment,toman2006influence}, neural network architectures seem to be more capable of overcoming sparsity thanks to the generalization power of word embeddings. 

\begin{table*}[t]
\begin{center}
\setlength{\tabcolsep}{7.0pt}
\scalebox{0.91}{
      \begin{tabular}{c l c  c  c  c c c c c c}
        \toprule
  \multicolumn{1}{c}{} &     
  \multirow{2}{*}{\small \textbf{\parbox{2cm}{Embedding Preprocessing}}}   &  \multicolumn{4}{c}{\textbf{Topic categorization}}  &  \multicolumn{5}{c}{\textbf{Polarity detection}}  \\

      \cmidrule(lr){3-6}
      \cmidrule(lr){7-11}
     
&     &   \multicolumn{1}{c}{\textbf{BBC}} & \multicolumn{1}{c}{\textbf{20News}} & \multicolumn{1}{c}{\textbf{Reuters}} & \multicolumn{1}{c}{\textbf{Ohsumed}} & \multicolumn{1}{c}{\textbf{RTC}} &
      \multicolumn{1}{c}{\textbf{IMDB}} &
      \multicolumn{1}{c}{\textbf{PL05}} &  \multicolumn{1}{c}{\textbf{PL04}} & 
      \multicolumn{1}{c}{\textbf{SF}}\\
      
    \midrule
    
    \multirow{4}{*}{\rotatebox[origin=c]{90}{\small \textbf{CNN}}}  &    \multicolumn{1}{|l}{\text{Vanilla}} &  94.6 & 89.2 & 93.7 & 35.3 & 83.2  & ~~87.5$^\dagger$ & \bf 76.3 & ~~58.7$^\dagger$ & \bf 91.2 \\
     &   \multicolumn{1}{|l}{\text{Lowercased}} & ~~93.9$^\dagger$ & ~~84.6$^\dagger$ & \bf 93.9 & \bf 36.2 & 83.2  & ~~85.4$^\dagger$ & \bf 76.3 & ~~60.0$^\dagger$ & 91.1 \\
    &   \multicolumn{1}{|l}{\text{Lemmatized}}  & 94.5  & ~~88.7$^\dagger$ & 93.8 & 35.4 & 83.0  & ~~86.8$^\dagger$ & 75.6 & 62.5 & \bf 91.2  \\ 
    &   \multicolumn{1}{|l}{\text{Multiword}}  & \bf 95.6  & \bf 89.7 & \bf 93.9 & 35.2 & \bf 83.3 & \bf 88.1 & 75.9 & \bf 63.1 & \bf 91.2  \\
        \midrule
       \midrule



        \multirow{4}{*}{\rotatebox[origin=c]{90}{\small \textbf{CNN+LSTM}}}  &    \multicolumn{1}{|l}{\text{Vanilla}} & 97.0  & ~~90.7$^\dagger$ & \bf 93.1 & ~~30.8$^\dagger$ & \bf 84.8  & \bf 88.9 & 79.1 & 71.4 & 87.1$^\dagger$ \\
      &  \multicolumn{1}{|l}{\text{Lowercased}}& 96.4  & \bf 91.8 & ~~92.5$^\dagger$ & ~~30.2$^\dagger$ & 84.5  & ~~88.0$^\dagger$ & 79.0 & \bf 74.2 & 87.4  \\
     \multicolumn{1}{l}{\textbf{}}  &   \multicolumn{1}{|l}{\text{Lemmatized}} & 96.6  & 91.5 & ~~92.5$^\dagger$ & ~~31.7$^\dagger$ & 83.9  & ~~86.6$^\dagger$ & ~~78.4$^\dagger$ & ~~67.7$^\dagger$ & 87.3 \\ 
    \multicolumn{1}{c}{\textbf{}}  &    \multicolumn{1}{|l}{\text{Multiword}} & \bf 97.3  & 91.3 &  92.8 & \bf 33.6 & 84.3  & ~~87.3$^\dagger$ & \bf 79.5 & 71.8 & \bf 87.5  \\

       
      

       

       \bottomrule
       
    \end{tabular}
    }
    \end{center}
    \caption{Cross-preprocessing evaluation: accuracy on the topic categorization and polarity detection tasks using different sets of word embeddings to initialize the embedding layer of the two classifiers. All datasets were preprocessed similarly according to the vanilla setting. $^\dagger$ indicates results that are statistically significant with respect to the top result. 
    }
    \label{tab:results-cross}
\end{table*}




\subsection{Experiment 2: Cross-preprocessing}
\label{experiment2}

This experiment aims at studying the impact of using different word embeddings (with differently preprocessed training corpora) on tokenized datasets (vanilla setting). 
Table \ref{tab:results-cross} shows the results for this experiment. 
In this experiment we observe a different trend, with multiword-enhanced vectors exhibiting a better performance both on the single CNN model (best overall performance in seven of the nine datasets) and on the CNN+LSTM model (best performance in four datasets and in the same ballpark as the best results in four of the remaining five datasets).
In this case the same set of words is learnt but single tokens inside multiword expressions are not trained. Instead, these single tokens are considered in isolation only, without the added \textit{noise} when considered inside the multiword expression as well. For instance, the word \textit{Apple} has a clearly different meaning in isolation from the one inside the multiword expression \textit{Big\_Apple}, hence it can be seen as beneficial not to train the word \textit{Apple} when part of this multiword expression. Interestingly, using multiword-wise embeddings on the vanilla setting leads to consistently better results than using them on the same multiword-grouped preprocessed dataset in eight of the nine datasets. This could provide hints on the excellent results provided by pre-trained Word2vec embeddings trained on the Google News corpus, which learns multiwords similarly to our setting.

Apart from this somewhat surprising finding, the use of the embeddings trained on a simple tokenized corpus (i.e. vanilla) proved again competitive, as different preprocessing techniques such as lowercasing and lemmatizing do not seem to help. In fact, the relatively weaker performance of lemmatization and lowercasing in this cross-processing experiment is somehow expected as the coverage of word embeddings in vanilla-tokenized datasets is limited, e.g., many entities which are capitalized in the datasets are not covered in the case of lowercasing, and inflected forms are missing in the case of lemmatizing.  



\section{Conclusions}

In this paper we analyzed the impact of simple text preprocessing decisions on the performance of a standard word-based neural text classifier. Our evaluations highlight the importance of being careful in the choice of how to preprocess our data and to be consistent when comparing different systems. In general, a simple tokenization works equally or better than more complex preprocessing techniques such as lemmatization or multiword grouping, except for domain-specific datasets (such as the medical dataset in our experiments) in which sole tokenization performs poorly. Additionally, word embeddings trained on multiword-grouped corpora perform surprisingly well when applied to simple tokenized datasets. This property has often been overlooked and, to the best of our knowledge, we test the hypothesis for the first time. 
In fact, this finding could partially explain the long-lasting success of pre-trained Word2vec embeddings, which specifically learn multiword embeddings as part of their pipeline \cite{mikolov2013distributed}. 

Moreover, our analysis shows that there is a high variance in the results depending on the preprocessing choice ($\pm 2.4\%$ on average for the best performing model), especially when the training data is not large enough to generalize. Further analysis and experimentation would be required to fully understand the significance of these results; but, this work can be viewed as a starting point for studying the impact of text preprocessing in deep learning models. We hope that our findings will encourage future researchers to carefully select and report these preprocessing decisions when evaluating or comparing different models. Finally, as future work, we plan to extend our analysis to other tasks (e.g. question answering), languages (particularly morphologically rich languages for which these results may vary
) and preprocessing techniques (e.g. stopword removal or part-of-speech tagging). 



\section*{Acknowledgments}

Jose Camacho-Collados is supported by the ERC Starting Grant 637277.




\bibliography{emnlp2018}
\bibliographystyle{acl_natbib_nourl}

\appendix



\end{document}